\pdfoutput=1
\documentclass[11pt]{article}

\usepackage[]{acl}

\usepackage{times}
\usepackage{latexsym}
\usepackage{cancel}
\usepackage{graphicx}
\usepackage{amsmath}
\usepackage{amsfonts}
\usepackage{amssymb}
\usepackage{bbm}
\usepackage{subcaption}
\usepackage{cleveref}

\usepackage{booktabs}
\usepackage{multirow}

\usepackage[T1]{fontenc}

\usepackage[utf8]{inputenc}

\usepackage{microtype}
\usepackage{adjustbox}
\usepackage[colorinlistoftodos,prependcaption,textsize=small]{todonotes}
\definecolor{darkpink}{rgb}{0.91, 0.33, 0.5}
\definecolor{lightpastelpurple}{rgb}{0.69, 0.61, 0.85}
\definecolor{darkpastelgreen}{rgb}{0.01, 0.75, 0.24}

\definecolor{Gray}{gray}{0.9}
\definecolor{bottlegreen}{rgb}{0.0,0.42,0.31}

\title{
        Don’t Forget About Pronouns: Removing Gender Bias in Language Models Without Losing Factual Gender Information}

\author{Tomasz Limisiewicz \and David Mare\v{c}ek \\
    Institute of Formal and Applied Linguistics, Faculty of Mathematics and Physics \\
    Charles University, Prague, Czech Republic \\
  \texttt{\{limisiewicz, marecek\}@ufal.mff.cuni.cz}
}

\begin{document}
\maketitle
\begin{abstract}The representations in large language models contain multiple types of gender information. We focus on two types of such signals in English texts: factual gender information, which is a grammatical or semantic property, and gender bias, which is the correlation between a word and specific gender. We can disentangle the model’s embeddings and identify components encoding both types of information with probing. We aim to diminish the stereotypical bias in the representations while preserving the factual gender signal. Our filtering method shows that it is possible to decrease the bias of gender-neutral profession names without significant deterioration of language modeling capabilities. The findings can be applied to language generation to mitigate reliance on stereotypes while preserving gender agreement in coreferences.\footnote{Our code is available on GitHub: \url{github.com/tomlimi/Gender-Bias-vs-Information}}
\end{abstract}

\section{Introduction}
Neural networks are successfully applied in natural language processing. While they achieve state-of-the-art results on various tasks, their decision process is not yet fully explained \cite{lipton_mythos_2018}. It is often the case that neural networks base their prediction on spurious correlations learned from large uncurated datasets. An example of such a spurious tendency is gender bias. Even the state-of-the-art models tend to counterfactually associate some words with a specific gender~\cite{zhao-etal-2018-gender, stanovsky-etal-2019-evaluating}. The representations of profession names tend to be closely connected with the stereotypical gender of their holders. When the model encounters the word ``nurse'', it will tend to use female pronouns (``she'', ``her'') when referring to this person in the generated text. This tendency is reversed for words such as ``doctor'', ``professor'', or ``programmer'',  which are male-biased.

\begin{figure}[t]
    \centering
    \includegraphics[width=\linewidth]{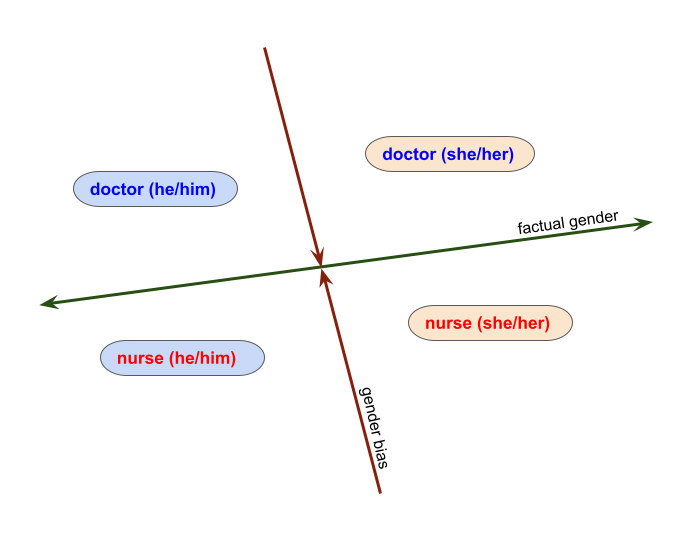}
     \caption{A schema is presenting the distinction between gender bias of nouns and factual (i.e., grammatical)  gender in pronouns. We want to transform the representations to mitigate the former and preserve the latter.
     }
    \label{fig:biase-vs-grammatical}

\end{figure}

It means that the neural model is not reliable enough to be applied in high-stakes language processing tasks such as connecting job offers to applicants' CVs \cite{de-arteaga-2019-bias}. If the underlying model was biased, the high-paying jobs, which are stereotypically associated with men, could be inaccessible for female candidates. When we decide to use language models for that purpose, the key challenge is to ensure that their predictions  are fair.

The recent works on the topics aimed to diminish the role of gender bias by feeding examples of unbiased text and training the network \cite{de-vassimon-manela-etal-2021-stereotype} or transforming the representations of the neural networks post-hoc (without additional training) \cite{bolukbasi-etal-man}. However, those works relied on the notion that to de-bias representation, most gender signal needs to be eliminated. It is not always the case, pronouns and a few other words (e.g.:``king'' -``queen''; ``boy'' - ``girl'') have factual information about gender.
A few works identified gendered words and exempted them from de-biasing \cite{zhao-etal-2018-learning, kaneko-bollegala-2019-gender}. In contrast to these approaches, we focus on contextual word embeddings. In contextual representations, we want to preserve the factual gender information for gender-neutral words when it is indicated by context, e.g., personal pronoun. This sort of information needs to be maintained in the representations. In language modeling, the network needs to be consistent about the gender of a person if it was revealed earlier in the text. The model’s ability to encode factual gender information is crucial for that purpose.

We propose a method for disentangling the factual gender information and gender bias encoded in the representations. We hypothesise that semantic gender information (from pronouns) is encoded in the network distinctly from the stereotypical bias of gender-neutral words (Figure~\ref{fig:biase-vs-grammatical}). We apply an orthogonal probe, which proved to be useful, e.g., in separating lexical and syntactic information encoded in the neural model \cite{limisiewicz-marecek-2021-introducing}. Then we filter out the bias subspace from the embedding space and keep the subspace encoding factual gender information. We show that this method performs well in both desired properties: decreasing the network’s reliance on bias while retaining knowledge about factual gender.



\subsection{Terminology}

We consider two types of gender information encoded in text:

\begin{itemize}
    \item \textbf{Factual gender} is the grammatical (pronouns ``he'', ``she'', ``her'', etc.) or semantic (``boy'', ``girl'', etc.) feature of specific word. It can also be indicated by a coreference link. We will call words with factual gender as \emph{gendered} in contrast to \emph{gender-neutral} words.
    \item \textbf{Gender bias} is the connection between a word and the specific gender with which it is usually associated, regardless of the factual premise.\footnote{For instance, the words ``nurse'', ``housekeeper'' are associated with women, and words ``doctor'', ``mechanic'' with men. None of those words has a grammatical gender marking in English.} \ We will refer to words with gender bias as \emph{biased} in contrast to \emph{non-biased}. 
\end{itemize}

Please note that those definitions do not preclude the existence of biased and at the same time gender-neutral words. In that case, we consider bias stereotypical and aim to mitigate it in our method. On the other hand, we want to preserve bias in gendered words.

\section{Methods}
We aim to remove the influence of gender-biased words while keeping the information about factual gender in the sentence given by pronouns. We focus on interactions of gender bias and factual gender information in coreference cues of the following form:

\begin{center}
    \small
    \textcolor{bottlegreen}{[NOUN]} examined the farmer for injuries because \textcolor{bottlegreen}{[PRONOUN]} was caring.
\end{center}


In English, we can expect to obtain the factual gender from the pronoun. Revealing one of the words in coreference link should impact the prediction of the other. Therefore we can name two causal associations:

\begin{itemize}
    \centering
    \item[$C_{I}$:] $\text{bias}_\text{noun}$ $\rightarrow$ $\text{f. gender}_\text{pronoun}$
    \item[$C_{II}$:] $\text{f. gender}_\text{pronoun}$ $\rightarrow$ $\text{bias}_\text{noun}$
\end{itemize}

In our method, we will primarily focus on two ways bias and factual gender interact. For gender-neutral nouns (in association $C_I$), the effect on predicting masked pronouns would be primarily correlated with their gender bias. At the same time, the second association is desirable, as it reveals factual gender information and can improve the masked token prediction of a gendered word. We define two conditional probability distributions corresponding to those causal associations:


\begin{equation}
    \label{eqn:probabilities}
    \begin{split}
    P_{I}( y_{\text{pronoun}} | X, b) \\
    P_{II}( y_{\text{noun}} | X, f)
    \end{split}
\end{equation}

Where $y$ is a token predicted in the position of pronoun and noun, respectively; $X$ is the context for masked language modeling. $b$ and $f$ are bias and factual gender factors, respectively. We model the bias factor by using a gender-neutral biased noun. Below we present examples for introducing female and male bias: \footnote{We use [NOUN] and [PRONOUN] tokens for a better explanation, in practice, they both are masked by the same mask token, e.g. [MASK] in \textsc{BERT} \cite{devlin-etal-2019-bert}.}

\textbf{Example 1:}

\begin{itemize}
    \small
    \item[\textbf{$b_f$}] \textbf{\textcolor{red}{The nurse}} examined the farmer for injuries because \textcolor{bottlegreen}{[PRONOUN]} was caring.
    \item[\textbf{$b_m$}] \textbf{\textcolor{blue}{The doctor}} examined the farmer for injuries because \textcolor{bottlegreen}{[PRONOUN]} was caring
\end{itemize}

\newpage
Similarly, the factual gender factor is modeled by introducing a pronoun with a specific gender in the sentence:

\textbf{Example 2:}

\begin{itemize}
    \small
    \item[\textbf{$f_f$}] \textcolor{bottlegreen}{[NOUN]} examined the farmer for injuries because \textbf{\textcolor{red}{she}} was caring.
    \item[\textbf{$f_m$}] \textcolor{bottlegreen}{[NOUN]} examined the farmer for injuries because \textbf{\textcolor{blue}{he}} was caring.
\end{itemize}

We aim to diminish the role of bias in the prediction of pronouns of a specific gender. On the other hand, the gender indicated in pronouns can be useful in the prediction of a gendered noun. Mathematically speaking, we want to drop the conditionality on bias factor in $P_I$ from \cref{eqn:probabilities}, while keeping the conditionality on gender factor in $P_{II}$.

\begin{equation}
    \label{eqn:probabilities-2}
    \begin{split}
    P_{I}( y_{\text{pronoun}} | X, b) \rightarrow P_{I}( y_{\text{pronoun}} | X ) \\
    P_{II}( y_{\text{noun}} | X, f) \not\rightarrow P_{II}( y_{\text{noun}} | X )
    \end{split}
\end{equation}


To decrease the effect of gender signal from the words other than pronoun and noun, we introduce a baseline, where both pronoun and noun tokens are masked:

\textbf{Example 3:}

\begin{itemize}
    \small
    \item[\textbf{$\varnothing$}] \textcolor{bottlegreen}{[NOUN]} examined the farmer for injuries because \textcolor{bottlegreen}{[PRONOUN]} was caring.
\end{itemize}

\subsection{Evaluation of Bias}


Manifestation of gender bias may vary significantly from model to model and can be attributed mainly to the choice of the pre-training corpora as well as the training regime. 
We define \emph{gender preference} in a sentence by the ratio between the probability of predicting male and female pronouns:

\begin{equation}
    GP(X) = \frac{P_{I}([\text{pronoun}_{m}]|X)}{P_{I}([\text{pronoun}_{f}]|X)}
\end{equation}

To estimate the gender bias of a profession name, we compare the gender preference in a sentence where the profession word is masked (example 3 from the previous paragraph) and not masked (example 1). We define \emph{relative gender preference}:

\begin{equation}
    \small
    \label{eqn:gender-bias}
    RGP_{\text{noun}} =  \log(GP(X_{\text{noun}})) - \log(GP(X_{\varnothing}))
\end{equation}

$X_{\text{noun}}$ denotes contexts in which the noun is revealed (example 1), and $X_{\varnothing}$ corresponds to example 3, where we mask both the noun and the pronoun.
Our approach focuses on the bias introduced by a noun, especially profession name. We subtract $\log(GP(X_{\varnothing}))$ to single out the bias contribution coming from the noun.\footnote{Other parts of speech may also introduce gender bias, e.g., the verb ``to work''. We note that our setting can be generalized to all words, but it is outside of the scope of this work.}
We use logarithm, so the results around zero would mean that revealing noun does not affect \emph{gender preference}.\footnote{The \emph{relative gender preference} was inspired by \emph{total effect} measure proposed by \citet{vig-etal-2020-causal-mediation}.}

\subsection{Disentangling Gender Signals with Orthogonal Probe}

To mitigate the influence of bias on the predictions \cref{eqn:probabilities-2}, we focus on the internal representations of the language model. We aim to inspect contextual representations of words and identify their parts that encode the causal associations $C_I$ and $C_{II}$.
For that purpose, we utilize \emph{orthogonal structural probes} proposed by \citet{limisiewicz-marecek-2021-introducing}. 

In structural probing, the embedding vectors are transformed in a way so that distances between pairs of the projected embeddings approximate a linguistic feature, e.g., distance in a dependency tree \cite{hewitt-manning-2019-structural}.
In our case, we want to approximate the gender information introduced by a gendered pronoun $f$ (factual) and gender-neutral noun $b$ (bias). The $f$ takes the values $-1$ for female pronouns and, $1$ for male ones, and $0$ for gender-neutral ``they''.
The $b$ is the relative gender preference (\cref{eqn:gender-bias}) for a specific noun ($b \equiv RGP_{\text{noun}}$).

Our orthogonal probe consists of three trainable components:
\begin{itemize}
    \item $O$: \emph{orthogonal transformation}, mapping representation to new coordinate system. 
    \item $SV$: \emph{scaling vector}, element-wise scaling of the dimensions in a new coordinate systems. We assume that dimensions that store probed information are associated with large scaling coefficients. 
    \item $i$: \emph{intercept} shifting the representation.
\end{itemize}

$O$ is a tunable orthogonal matrix of size $d_{\text{emb}} \times d_{\text{emb}}$, $SV$ and $i$ are tunable vectors of length $d_{\text{emb}}$, where $d_{\text{emb}}$ is the dimensionality of model's embeddings.
The probing losses are the following:
\begin{equation}
    \small
    \label{eqn:probing}
    \begin{split}
    L_I = \bigl| || {SV}_{I} \odot( O \cdot (h_{b,P} - h_{\varnothing, P})) - i_{I} ||_{d} - b \bigr| \\
    L_{II}  = \bigl| || {SV}_{II} \odot ( O \cdot (h_{f,N} - h_{\varnothing, N})) - i_{II} ||_{d} - f \bigr|,
    \end{split}
\end{equation}
where, $h_{b,P}$ is the vector representation of masked pronoun in example 1; $h_{f, N}$ is the vector representation of masked noun in example 2; vectors $h_{\varnothing, P}$ and $h_{\varnothing, N}$ are the representations of masked pronoun and noun respectively in baseline example 3.




To account for negative values of target factors ($b$ and $f$) in \cref{eqn:probing}, we generalize distance metric to negative values in the following way:

\begin{equation}
    \label{eqn:dual-metric}
    || \overrightarrow{v} ||_{d} = ||\max(\overrightarrow{0}, \overrightarrow{v})||_2 - ||\min(\overrightarrow{0}, \overrightarrow{v}) ||_2
\end{equation}

We jointly probe for both objectives (orthogonal transformation is shared).
\citet{limisiewicz-marecek-2021-introducing} observed that the resulting scaling vector after optimization tends to be sparse, and thus they allow to find the subspace of the embedding space that encodes particular information.

\subsection{Filtering Algorithm}

In our algorithm we aim to filter out the latent vector's dimensions that encode bias. Particularly, we assume that, when $||h_{b,P} - h_{\varnothing, P}|| \rightarrow 0$ then $P_{I}( y_{\text{pronoun}} | X, b) \rightarrow P_{I}( y_{\text{pronoun}} | X) $

We can diminish the information by masking the dimensions with a corresponding scaling vector coefficient larger than small $\epsilon$.\footnote{We take epsilon equal to $10^{-12}$. Our results weren't particularly vulnerable to this parameter, we show the analysis in \cref{sec:filtering-threshold}.} 
The bias filter is defined as:
\begin{equation}
    \label{eqn:bias-filter}
    F_{-b} = \overrightarrow{\mathbbm{1}}[\epsilon > abs(SV_I)],
\end{equation}
where $abs(\cdot)$ is element-wise absolute value and $\overrightarrow{\mathbbm{1}}$ is element-wise indicator. We apply this vector to the representations of hidden layers:
\begin{equation}
    \label{eqn:filtering}
    \hat{h} = O^{T} \cdot ( F_{-b} \odot (O \cdot h) + abs(SV_I)
    \odot i_{I})
\end{equation}

To preserve factual gender information, we propose an alternative version of the filter. The dimension is kept when its importance (measured by the absolute value of scaling vector coefficient) is higher in probing for factual gender than in probing for bias. We define factual gender preserving filter as:
\begin{equation}
    \label{eqn:bias-gender-filter}
    F_{-b,+f} = F_{-b} + \overrightarrow{\mathbbm{1}}[\epsilon  \leq abs(SV_I) < abs(SV_{II})]
\end{equation}

The filtering is performed as in \cref{eqn:filtering}
We analyze the number of overlapping dimensions in two scaling vectors in Section \ref{sec:filter-analysis}.

\section{Experiments and Results}
\begin{table*}[t]
\centering
\begin{tabular}{@{}|l|c||c|@{}}
    \toprule
    Prompt    & PRONOUN & PRONOUN 2        \\ \midrule
    \textcolor{bottlegreen}{[PRONOUN]} is \textcolor{bottlegreen}{[NOUN]}.                                 & She He &       \\
    \textcolor{bottlegreen}{[PRONOUN]} was \textcolor{bottlegreen}{[NOUN]}.                                & She He &       \\
    \textcolor{bottlegreen}{[PRONOUN]}works as \textcolor{bottlegreen}{[NOUN]}.                           & She He &       \\
    \textcolor{bottlegreen}{[PRONOUN]} job is \textcolor{bottlegreen}{[NOUN]}.                             & Her His &     \\
    \textcolor{bottlegreen}{[NOUN]}said that \textcolor{bottlegreen}{[PRONOUN]} loves \textcolor{bottlegreen}{[PRONOUN 2]} job.    & he she &  her his \\
    \textcolor{bottlegreen}{[NOUN]} said that \textcolor{bottlegreen}{[PRONOUN]} hates \textcolor{bottlegreen}{[PRONOUN 2]} job.     & she he & her his \\ 
    \bottomrule
    \end{tabular}
\caption{List of evaluation prompts used in the evaluation of \emph{relative gender preference}. The tag [NOUN] masks a noun accompanied by an appropriate determiner.}
\label{tab:prompts}
\end{table*}

We examine the representation of two \textsc{BERT} models (base-cased: 12 layers, 768 embedding size; and large-cased: 24 layers, 1024 embedding size, \citet{devlin-etal-2019-bert}), and \textsc{ELECTRA} (base-generator: 12 layers, 256 embedding size \citet{clark2020electra}). All the models are Transformer encoders trained on the masked language modeling objective.

\subsection{Evaluation of Gender Bias in Language Models}

Before constructing a de-biasing algorithm, we evaluate the bias in the prediction of three language models.

We evaluate the gender bias in language models on 104 gender-neutral professional words from the WinoBias dataset \cite{zhao-etal-2018-gender}. The authors analyzed the data from the US Labor Force Statistics. They annotated 20 professions with the highest share of women as stereotypically female and 20 professions with the highest share of men as stereotypically male.

We run the inference on the prompts in five formats presented in Table~\ref{tab:prompts} and estimate with equation \cref{eqn:gender-bias}. To obtain the bias of the word in the model, we take mean $RGP_{noun}$ computed on all prompts.



\subsubsection{Results}

\begin{table*}[t]
\centering
\begin{tabular}{@{}l|ccc||l|ccc@{}}
\toprule
\multicolumn{4}{c||}{Most Female Biased}       & \multicolumn{4}{c}{Most Male Biased}        \\ \midrule
NOUN & N Models & Avg. RGP & Annotated & NOUN & N Models & Avg. RGP & Annotated \\ \midrule
housekeeper  & 3/3    & -2.009   & female    & carpenter   & 3/3    & 0.870    & male      \\
nurse        & 3/3    & -1.840   & female    & farmer      & 3/3    & 0.753    & male      \\
receptionist & 3/3    & -1.602   & female    & guard       & 3/3    & 0.738    & male      \\
hairdresser  & 3/3    & -0.471   & female    & sheriff     & 3/3    & 0.651    & male      \\
librarian    & 2/3    & -0.279   & female    & firefighter & 3/3    & 0.779    & \bf{neutral}   \\
victim       & 2/3    & -0.102   & \bf{neutral}   & driver      & 3/3    & 0.622    & male      \\
child        & 2/3    & -0.060   & \bf{neutral}   & mechanic    & 2/3    & 0.719    & male      \\
salesperson  & 2/3    & -0.056   & \bf{male}      & engineer    & 2/3    & 0.645    & \bf{neutral}   \\ \bottomrule
\end{tabular}
\caption{Evaluated empirical bias in analyzed Masked Language Models. Column number shows the count of models for which the word was considered biased. Annotated is the bias assigned in \citet{zhao-etal-2018-gender} based on the job market data.}
\label{tab:empirical-bias}
\end{table*}

We compare our results with the list of stereotypical words from the annotation of \citet{zhao-etal-2018-gender}. Similarly, we pick up to 20 nouns with the highest and positive $RGP$ as male-biased and up to 20 nouns with the lowest and negative $RGP$ as  female-biased. These lists differ for models.

Table~\ref{tab:empirical-bias} presents the most biased words according to three models. Noticeably, there are differences between empirical and annotated bias. Especially word ``salesperson'' considered male-biased based on job market data was one of the most skewed toward the female gender in 2 out of 3 models. The full results of the evaluation can be found in \cref{sec:bias-in-lms}.



\subsection{Probing for Gender Bias and Factual Gender Information}
\label{sec:filter-analysis}

We optimize the joint probe, where orthogonal transformation is shared, while scaling vectors and intercepts are task specific. The probing objective is to approximate: $\boldsymbol{C_{I}}$) the gender bias of gender-neutral nouns ($b \equiv RGP_{\text{noun}}$); and $\boldsymbol{C_{II}}$) the factual gender information of pronouns ($f \equiv \text{f. gender}_{\text{pronoun}}$). 

We use WinoMT dataset\footnote{The dataset was originally introduced to evaluate gender bias in machine translation.} 
\cite{stanovsky-etal-2019-evaluating} which is a derivate of WinoBias dataset \cite{zhao-etal-2018-gender}. Examples are more challenging to solve in this dataset than in our evaluation prompts (Table~\ref{tab:prompts}). Each sentence contains two potential antecedents. We use WinoMT for probing because we want to separate probe optimization and evaluation data. 
Moreover, we want to identify the encoding of gender bias and factual gender information in more diverse contexts.

We split the dataset into train, development, and test sets with non-overlapping nouns, mainly profession names. They contain 62, 21, and 21 unique nouns, corresponding to 2474, 856, and 546 sentences. The splits are designed to balance male and female-biased words in each of them.

\subsubsection{Results}

\begin{figure*}[t]
    \begin{subfigure}{0.27\textwidth}
        \centering
        \includegraphics[width=\textwidth]{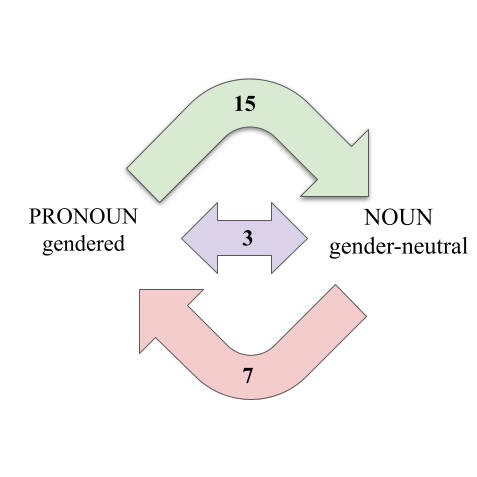}
        \caption{\textsc{BERT} base (out of 768 dims)}
    \end{subfigure}
    \hfill
    \begin{subfigure}{0.27\textwidth}
        \centering
        \includegraphics[width=\textwidth]{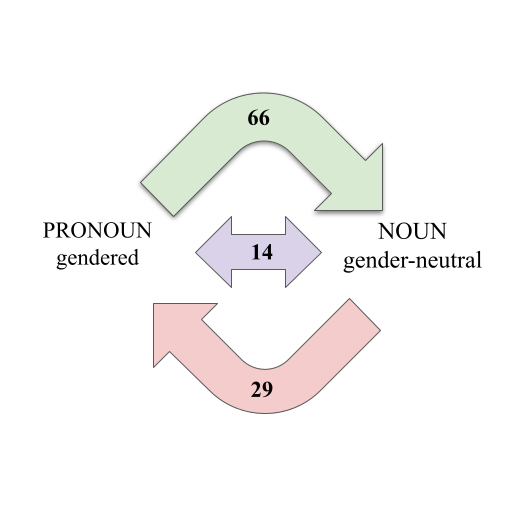}
        \caption{\textsc{BERT} large (out of 1024 dims)}
    \label{fig:doc2}
    \end{subfigure}
    \hfill
    \begin{subfigure}{0.27\textwidth}
        \centering
        \includegraphics[width=\textwidth]{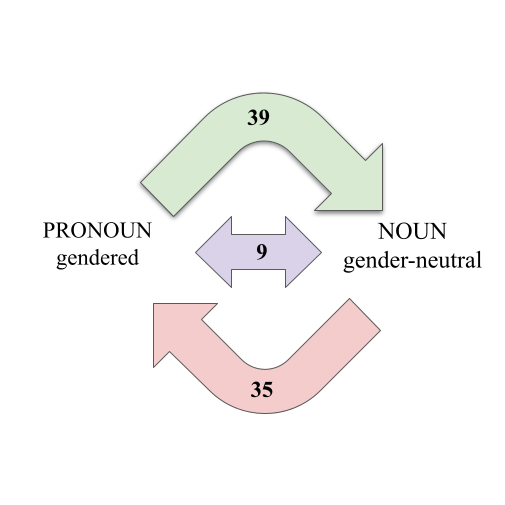}
        \caption{\textsc{ELECTRA} (out of 256 dims)}
    \end{subfigure}
    \caption{The number of selected dimensions for each of the tasks: \textcolor{darkpink}{$\boldsymbol{C_{I}}$}, \textcolor{darkpastelgreen}{$\boldsymbol{C_{II}}$}, and \textcolor{lightpastelpurple}{shared for both tasks}.}
    \label{fig:selected-dimensions}
\end{figure*}

The probes on the models' top layer give a good approximation of factual gender -- Pearson correlation between predicted and gold values in the range from $0.928$ to $0.946$. Pearson correlation for bias was high for \textsc{BERT} base ($0.876$), \textsc{BERT} large ($0.946$),  and lower for \textsc{ELECTRA} ($0.451$).\footnote{For \textsc{ELECTRA}, we observed higher correlation of the bias probe on penultimate layer $0.668$.}

We have identified the dimensions encoding conditionality $\boldsymbol{C_{I}}$ and $\boldsymbol{C_{II}}$. In Figure~\ref{fig:selected-dimensions}, we present the number of dimensions selected for each objective and their overlap. We see that bias is encoded sparsely in 18  to 80 dimensions.

\subsection{Filtering Gender Bias}
\label{sec:bias-filtering-results}

The primary purpose of probing is to construct bias filters based on the values of scaling: $F_{-b}$ and $F_{-b,+f}$. Subsequently, we perform our de-biasing transformation \cref{eqn:bias-filter} on the last layers of the model. The probes on top of each layer are optimized separately.


After filtering, we again compute $RGP$ for all professions. We monitor the following metrics to measure the overall improvement of the de-biasing algorithm on the  set of 104 gender-neutral nouns $S_{GN}$:

\begin{equation}
    \label{eqn:mse-gender-neutral}
    MSE_{GN} = \frac{1}{|S_{GN}|}\sum_{w \in S_{GN}} RGP(w)^2
\end{equation}

\emph{Mean squared error} show how far from zero $RGP$ is. The advantage of this metric is that the bias of some words cannot be compensated by the opposite bias of others. The main objective of de-biasing is to minimize mean squared error. 

\begin{equation}
    \label{eqn:mean-gender-neutral}
    MEAN_{GN} = \frac{1}{|S_{GN}|}\sum_{w \in S_{GN}} RGP(w)^2
\end{equation}

Mean shows whether the model is skewed toward predicting specific gender. In cases when the mean is close to zero, but $MSE$ is high, we can tell that there is no general preference of the model toward one gender, but the individual words are biased.

\begin{equation}
    \label{eqn:var-gender-neutral}
    VAR_{GN} = MSE_{GN} - MEAN_{GN}^2
\end{equation}

Variance is a similar measure to $MSE$. It is useful to show the spread of $RGP$ when the mean is non-zero.

Additionally, we introduce a set of 26 gendered nouns ($S_{G}$) for which we expect to observe non-zero $RGP$. We monitor $MSE$ to diagnose whether semantic gender information is preserved in de-biasing:

\begin{equation}
    \label{eqn:mse-gendered}
    MSE_{G} = \frac{1}{|S_{G}|}\sum_{w \in S_{G}} RGP(w)
\end{equation}

\subsubsection{Results}

\begin{table}[t]
\small
\begin{adjustbox}{width=\columnwidth,center}
\begin{tabular}{l|c|c|ccc}
\toprule
\multirow{2}{*}{Setting} & \multirow{2}{*}{FL} & $MSE$ & $MSE$ & $MEAN$ & $VAR$    \\ \cline{3-6} 
    &   & gendered & \multicolumn{3}{c}{gender-neutral} \\ \midrule
BERT B    & - & \bf6.177 & 0.504 & 0.352 & 0.124  \\
-bias         & 1 & 2.914 & 0.136 & \bf-0.056 & 0.133 \\
              & 2 & 2.213 & \bf0.102 & -0.121 & \bf0.088 \\ 
+f. gender & 1 & 3.780 & 0.184 & -0.067 & 0.180 \\
              & 2 & 2.965 & 0.145 & -0.144 & 0.124 \\ \midrule \midrule
ELECTRA       & - & \bf1.360 & 0.367 & 0.163 & 0.340  \\ 
-bias         & 1 & 0.100 & 0.124 & 0.265 & 0.054  \\
              & 2 & 0.048 & \bf0.073 & 0.200 & \bf0.033  \\
+f. gender & 1 & 0.901 & 0.186 & \bf0.008 & 0.185  \\
              & 2 & 0.488 & 0.101 & -0.090 & 0.093 \\ \midrule \midrule
BERT L    & - & \bf1.363 & 0.099 & 0.235  & 0.044 \\
-bias         & 1 & 0.701 & 0.051 & 0.166 & 0.024  \\
              & 2 & 0.267 & 0.015 & 0.069 & 0.011  \\
              & 4 & 0.061 & 0.033 & 0.162 & \bf0.007 \\
+f. gender & 1 & 1.156 & 0.057 & 0.145 & 0.036  \\
              & 2 & 0.755 & 0.020 & \bf0.011 & 0.020  \\
              & 4 & 0.292 & \bf0.010 & 0.037 & 0.009  \\ \midrule
\multicolumn{2}{c|}{\textbf{AIM:}}  & $\uparrow$ & $\downarrow$ & $\approx 0$ & $\downarrow$ \\
\bottomrule
\end{tabular}
\end{adjustbox}
\caption{Aggregation of \emph{relative gender preference} in prompts for gendered and gender-neutral nouns. FL denotes the number of the model's top layers for which filtering was performed.}
\label{tab:rgp-results}
\end{table}

In Table~\ref{tab:rgp-results}, we observe that in all cases, gender bias measured by $MSE_{GN}$ decreases after filtering of bias subspace. The filtering on more than one layer usually further brings this metric down. It is important to note that the original model differs in the extent to which their predictions are biased. The mean square error is the lowest for \textsc{BERT} large ($0.099$), noticeably it is lower than in other analyzed models after de-biasing (except for \textsc{ELECTRA} after 2-layer filtering $0.073$).

The predictions of all the models are skewed toward predicting male pronoun when the noun is revealed. 
Most of the pronouns used in the evaluation were professional names. Therefore, we think that this result is the manifestation of the stereotype that career-related words tend to be associated with men.

After filtering \textsc{BERT} base becomes slightly skewed toward female pronouns ($MEAN_{GN} < 0$). For the two remaining models, we observe that keeping factual gender signal performs well in decreasing $MEAN_{GN}$.

Another advantage of keeping factual gender representation is the preservation of the bias in semantically gendered nouns, i.e., higher $MSE_{G}$.

\subsection{How Does Bias Filtering Affect Masked Language Modeling?}
\label{sec:MLM-effect}

We examine whether filtering affects the model's performance on the original task. For that purpose, we evaluate top-1 prediction accuracy for the masked tokens in the test set from English Web Treebank UD \cite{silveira14gold} with 2077 sentences.
We also evaluate the capability of the model to infer the personal pronoun based on the context. We use the GAP Coreference Dataset \cite{webster2018gap} with 8908 paragraphs. In each test case, we mask a pronoun referring to a person usually mentioned by their name. In the sentences, gender can be easily inferred from the name. In some cases, the texts also contain other (un-masked) gender pronouns.

\subsubsection{Results: All Tokens}

The results in Table~\ref{tab:mlm-ewt} show that filtering out bias dimensions moderately decrease MLM accuracy: up to $0.037$ for \textsc{BERT} large; $0.052$ for \textsc{BERT} base; $0.07$ for \textsc{ELECTRA}. In most cases exempting factual gender information from filtering decreases the drop in results.

\begin{table}[t]
\small
\centering
\begin{tabular}{@{}l|c|ccc@{}}
\toprule
\multirow{2}{*}{Setting} & \multirow{2}{*}{FL} & \multicolumn{3}{c}{Accuracy}     \\ \cline{3-5}
 & & BERT L & BERT B & ELECTRA \\ \midrule

Original & - & 0.516 & 0.526 & 0.499 \\ 
-bias & 1 & 0.515 & 0.479 & 0.429 \\ 
 & 2 & 0.504 & 0.474 & 0.434 \\ 
 & 4 & 0.479 & -  & - \\ 
+f. gender & 1 & 0.515 & 0.479 & 0.434 \\ 
 & 2 & 0.510 & 0.480 & 0.433 \\ 
 & 4 & 0.489 & - & - \\
 \bottomrule
\end{tabular}
\caption{Top-1 accuracy for all tokens in EWT UD \cite{silveira14gold}. FT is the number of the model’s top layers for which filtering was performed.}
\label{tab:mlm-ewt}
\end{table}

\subsubsection{Results: Personal Pronouns in GAP}

We observe a more significant drop in results in the GAP dataset after de-biasing. The deterioration can be alleviated by omitting factual gender dimensions in the filter. For \textsc{BERT} large and \textsc{ELECTRA} this setting can even bring improvement over the original model. Our explanation of this phenomenon is that filtering can decrease the confounding information from stereotypically biased words that affect the prediction of correct gender.

In this experiment, we also examine the filter, which removes all factual-gender dimensions. Expectedly such a transformation significantly decreases the accuracy. However, we still obtain relatively good results, i.e., accuracy higher than $0.5$, which is a high benchmark for choosing gender by random. Thus, we conjecture that the gender signal is still left in the model despite filtering.

\begin{table}[t]
\centering
\begin{tabular}{@{}l|c|ccc@{}}
\toprule
\multirow{2}{*}{Setting}                      & \multirow{2}{*}{FL} & \multicolumn{3}{c}{Accuracy} \\ \cline{3-5}
 & & Overall & Male  & Female \\ \midrule
BERT L & - & 0.799 & 0.\bf816 & 0.781 \\ 
-bias  & 1  & 0.690  & 0.757  & 0.624 \\ 
  & 2  & 0.774  & 0.804  & 0.744 \\ 
  & 4  & 0.747  & 0.770  & 0.724 \\ 
+f. gender  & 1  & 0.754  & 0.782  & 0.726 \\ 
  & 2  & 0.785  & 0.801  & 0.769 \\ 
  & 4  & \bf0.801  & 0.807  & \bf0.794 \\ \midrule 
-f. gender  & 1  & 0.725  & 0.775  & 0.675 \\ 
  & 2  & 0.763  & 0.788  & 0.738 \\ 
  & 4  & 0.545  & 0.633  & 0.458 \\ \midrule \midrule
BERT B  & -  & \bf0.732  & \bf0.752  & \bf0.712 \\ 
-bias  & 1  & 0.632  & 0.733  & 0.531 \\ 
  & 2  & 0.597  & 0.706  & 0.487 \\ 
+f. gender  & 1  & 0.659  & 0.734  & 0.584 \\ 
  & 2  & 0.620  & 0.690  & 0.549 \\ \midrule
-f. gender  & 1  & 0.634  & 0.662  & 0.606 \\ 
  & 2  & 0.604  & 0.641  & 0.567 \\ \midrule \midrule
ELECTRA  & -  & 0.652  & 0.680  & 0.624 \\ 
-bias  & 1  & 0.506  & 0.731  & 0.280 \\ 
  & 2  & 0.485  & 0.721  & 0.249 \\ 
+f. gender  & 1  & \bf0.700  & \bf0.757  & \bf0.642 \\ 
  & 2  & 0.691  & 0.721  & 0.661 \\  \midrule
-f. gender  & 1  & 0.395  & 0.660  & 0.129 \\ 
  & 2  & 0.473  & 0.708  & 0.239 \\ 
\bottomrule
\end{tabular}
\caption{Top-1 accuracy for masked pronouns in GAP dataset \cite{webster2018gap}. FT is the number of the model’s top layers for which filtering was performed.}
\label{tab:mlm-gap}
\end{table}

\paragraph{Summary of the Results:} We observe that the optimal de-biasing setting is factual gender preserving filtering ($F_{-b,+f}$). This approach diminishes stereotypical bias in nouns while preserving gender information for gendered nouns (\cref{sec:bias-filtering-results}). Moreover, it performs better in masked language modeling tasks (\cref{sec:MLM-effect}).


\section{Related Work}
In recent years, much focus was put on evaluating and countering bias in language representations or word embeddings. \citet{bolukbasi-etal-man} observed the distribution of Word2Vec embeddings \citep{mikolov-2013-w2v} encode gender bias. They tried to diminish its role by projecting the embeddings along the so-called \textit{gender direction}, which separates gendered words such as \textit{he} and \textit{she}. They measure the bias as cosine similarity between an embedding and the gender direction.

\begin{equation}
    \text{GenderDirection} \approx \overrightarrow{he} - \overrightarrow{she}
\end{equation}

\citet{zhao-etal-2018-learning} propose a method to diminish differentiation of word representations in the gender dimension during training of the GloVe embeddings \citep{pennington-etal-2014-glove}. Nevertheless, the following analysis of \citet{gonen-goldberg-2019-lipstick} argued that these approaches remove bias only partially and showed that bias is encoded in the multi-dimensional subspace of the embedding space. The issue can be resolved by projecting in multiple dimensions to further nullify the role of gender in the representations \cite{ravfogel-etal-2020-null}. Dropping all the gender-related information, e.g., the distinction between feminine and masculine pronouns can be detrimental to gender-sensitive applications. \citet{kaneko-bollegala-2019-gender} proposed a de-biasing algorithm that preserves gendered information in gendered words.

Unlike the approaches above, we work with contextual embeddings of language models. \citet{vig-etal-2020-causal-mediation} investigated bias in the representation of the contextual model (GPT-2, \citet{radford2019language}). They used causal mediation analysis to identify components of the model responsible for encoding bias. \citet{nadeem-etal-2021-stereoset} and \citet{nangia-etal-2020-crows} propose a method of evaluating bias (including gender) with counterfactual test examples, to some extent similar to our prompts.

\citet{qian-etal-2019-reducing} and \citet{liang-etal-2020-bert} employ prompts similar to ours to evaluate the gender bias of professional words in language models. The latter work also aims to identify and remove gender subspace in the model. In contrast to our approach, they do not guard factual gender signal.

Recently, \citet{stanczak2021survey} summarized the research on the evaluation and mitigation of gender bias in the survey of 304 papers.








\section{Discussion}
\subsection{Bias Statement}

We define bias as the connection between a word and the specific gender it is usually associated with. The association usually stems from the imbalanced number of corpora mentions of the word in male and female contexts. This work focuses on the stereotypical bias of nouns that do not have otherwise denotation of gender (semantic or grammatical). We consider such a denotation as factual gender and want to guard it in the models' representation. 

Our method is applied to language models, hence we recognize potential application in language generation. We envision the case where the language model is applied to complete the text about a person, where we don't have implicit information about their gender. In this scenario, the model should not be compelled by stereotypical bias to assign a specific gender to a person. On the other hand, when the implicit information about a person's gender is provided in the context, the generated text should be consistent.

Language generation is becoming  ubiquitous  in everyday NLP applications (e.g., chat-bots, auto-completion \citet{robert-2020-nlg}). Therefore it is important to ensure that the language models do not propagate sex-based discrimination. 

The proposed method can also be implemented in deep models for other tasks, e.g., machine translation systems. In machine translation, bias is especially harmful when translating from English to languages that widely denote gender grammatically. In translation to such languages generation of gendered nouns tends to be made based on stereotypical gender roles instead of factual gender information provided in the source language \cite{stanovsky-etal-2019-evaluating}.

\subsection{Limitations}
It is important to note that we do not remove the whole of the gender information in our filtering method. Therefore, a downstream classifier could easily retrieve the factual gender of a person mentioned in a text, e.g., their CV.

This aspect makes our method not applicable to downstream tasks that use gender-biased data. For instance, in the task of predicting a profession based on a person's biography \cite{de-arteaga-2019-bias}, there are different  proportions of men and women among holders of specific professions. A classifier trained on de-biased but not de-gendered embeddings would learn to rely on gender property in its predictions.


Admittedly, in our results, we see that the proposed method based on \emph{orthogonal probes} does not fully remove gender bias from the representations \cref{sec:bias-filtering-results}. Even though our method typically identifies multiple dimensions encoding bias and factual gender information, there is no guarantee that all such dimensions will be filtered. Noticeably, the de-biased  \textsc{BERT} base still underperform off-the-shelf \textsc{BERT} large in terms of $MSE_{GN}$. The reason behind this particular method was its ability to disentangle the representation of two language signals, in our case: gender bias and factual gender information.

Lastly, the probe can only recreate linear transformation, while in a non-linear system such as Transformer, the signal can be encoded non-linearly. Therefore, even when we remove the whole bias subspace, the information can be recovered in the next layer of the model \cite{ravfogel-etal-2020-null}. It is also the reason why we decided to focus on the top layers of models.




\section{Conclusions}
We propose a new insight into gender information in contextual language representations. In de-biasing, we focus on the trade-off between removing stereotypical bias while preserving the semantic and grammatical information about the gender of a word from its context. Our evaluation of gender bias showed that three analyzed masked language models (\textsc{BERT} large, \textsc{BERT} based, and \textsc{ELECTRA}) are biased and skewed toward predicting male gender for profession names. To mitigate this issue, we disentangle stereotypical bias from factual gender information. Our filtering method can remove the former to some extent and preserve the latter. As a result, we decrease the bias in predictions of language models without significant deterioration of their performance in masked language modeling task.

\section*{Aknowlegments}

We thank anonymous reviewers and our colleagues: João Paulo de Souza Aires, Inbal Magar, and Yarden Tal, who read the previous versions of this work and provided helpful comments and suggestions for improvement. The work has been supported by grant 338521 of the Grant Agency of Charles University.

\bibliographystyle{acl_natbib}
\bibliography{custom,anthology}

\appendix

\section{Technical Details}

We use batches of size $10$. Optimization is conducted with Adam \cite{kingma-2014-adam} with initial learning rate $0.02$ and meta parameters: $\beta_1=0.9$, $\beta_2=0.999$, and $\epsilon=10^{-8}$. We use learning rate decay and an early-stopping mechanism with a decay factor $10$. The training is stopped after three consecutive epochs not resulting in the improvement of the validation loss learning rate. We clip each gradient's norm at $c=1.0$. The orthogonal penalty was set to $\lambda_O = 0.1$.

We implemented the network in TensorFlow 2 \cite{tensorflow-2015-whitepaper}. The code will be available on GitHub.

\subsection{Computing Infrastructure}

We optimized probes on a GPU core \textit{GeForce GTX 1080 Ti}. Training a probe on top of one layer of \textsc{BERT} large takes about 5 minutes.

\subsection{Number of Parameters in the Probe}

The number of the parameters in the probe depends on the model's embedding size $d_{\text{emb}}$. The \emph{orthogonal transformation} matrix consist of $d_{\text{emb}}^2$; both \emph{intercept} and \emph{scalling vector} have $d_{\text{emb}}$ parameters. Altogether, the size of the probe equals to $d_{\text{emb}}^2 + 4 \cdot d_{\text{emb}}$.

\section{Details about Datasets}

WinoMT is distributed under MIT license; EWT UD under Creative Commons 4.0 license; GAP under Apache 2.0 license.

\section{Results for Different Filtering Thresholds}
\label{sec:filtering-threshold}

\begin{table}[]
\small
\begin{tabular}{l|c|ccc}
\toprule
\multirow{2}{*}{Epsilon} & $MSE$ & $MSE$ & $MEAN$ & $VAR$    \\ \cline{2-5} 
    & gendered & \multicolumn{3}{c}{gender-neutral} \\ \midrule
$10^{-2}$ & 0.762 & 0.083 & 0.233 & 0.029 \\ 
$10^{-4}$ & 0.756 & 0.081 & 0.230 & 0.028 \\ 
$10^{-6}$ & 0.764 & 0.074 & 0.213 & 0.029 \\ 
$10^{-8}$ & 0.738 & 0.078 & 0.225 & 0.027 \\ 
$10^{-10}$ & 0.721 & 0.082 & 0.234 & 0.027 \\ 
$10^{-12}$ & 0.701 & 0.051 & 0.166 & 0.024 \\ 
$10^{-14}$ & 0.709 & 0.043 & 0.138 & 0.023 \\ 
$10^{-16}$ & 0.770 & 0.023 & 0.013 & 0.022 \\
\bottomrule
\end{tabular}
\caption{Tuning of filtering threshold $\epsilon$. Results for filtering bias in the last layer of \textsc{BERT} large.}
\label{tab:threshold-tuning}
\end{table}

In \cref{tab:threshold-tuning} we show how the choice of filtering threshold $\epsilon$ affects the results of our method for \textsc{BERT} large. We decided to pick the threshold equal to $10^{-12}$, as lowering it brought only minor improvement in $MSE_{GN}$.

\begin{table}[!t]
\small
\begin{adjustbox}{width=\columnwidth,center}
\centering
\begin{tabular}{@{}l|ccc|c@{}}
\toprule
\multirow{2}{*}{NOUN}   & \multicolumn{4}{c}{Relative Gender Preference} \\ \cline{2-5} 
    & \textsc{BERT} base &  \textsc{BERT} large & \textsc{ELECTRA}  &  Avg.   \\ \midrule
\multicolumn{5}{c}{Female Gendered} \\ \midrule
councilwoman & -4.262 & -2.050 & -0.832 & -2.381 \\
policewoman & -4.428 & -1.710 & -0.928 & -2.355 \\
princess & -3.486 & -1.598 & -1.734 & -2.273 \\ 
actress & -3.315 & -1.094 & -2.319 & -2.242 \\ 
chairwoman & -4.020 & -1.818 & -0.629 & -2.156 \\ 
waitress & -2.806 & -1.167 & -2.475 & -2.150 \\ 
busimesswoman & -3.202 & -1.696 & -1.096 & -1.998 \\ 
queen & -2.752 & -0.910 & -2.246 & -1.969 \\
spokeswoman & -2.543 & -2.126 & -1.017 & -1.895 \\ 
stewardess & -3.484 & -2.215 & 0.089 & -1.870 \\
maid & -3.092 & -0.822 & -1.452 & -1.788 \\ 
witch & -2.068 & -0.706 & -1.476 & -1.416 \\ 
nun & -2.472 & -0.974 & -0.613 & -1.353 \\ \bottomrule
\multicolumn{5}{c}{Male Gendered} \\ \midrule
wizard & 0.972 & 0.314 & 0.237 & 0.508 \\ 
manservant & 0.974 & 0.493 & 0.115 & 0.527 \\ 
steward & 0.737 & 0.495 & 0.675 & 0.636 \\ 
spokesman & 0.846 & 0.591 & 0.515 & 0.651 \\ 
waiter & 1.003 & 0.473 & 0.639 & 0.705 \\ 
priest & 0.988 & 0.442 & 0.928 & 0.786 \\ 
actor & 1.366 & 0.392 & 0.632 & 0.797 \\ 
prince & 1.401 & 0.776 & 0.418 & 0.865 \\ 
policeman & 1.068 & 0.514 & 1.202 & 0.928 \\ 
king & 1.399 & 0.658 & 0.772 & 0.943 \\ 
chairman & 1.140 & 0.677 & 1.069 & 0.962 \\ 
councilman & 1.609 & 1.040 & 0.419 & 1.023 \\ 
businessman & 1.829 & 0.549 & 0.985 & 1.121 \\ 
\bottomrule
\end{tabular}
\end{adjustbox}
\caption{List of gendered nouns with evaluated bias in three analyzed models ($RGP$).}
\label{tab:gendered-empirical-bias}
\end{table}

\section{Evaluation of Bias in Language Models}
\label{sec:bias-in-lms}

We present the list of 26 gendered words and their empirical bias in \cref{tab:gendered-empirical-bias}. Following tables \cref{tab:gender-neutral-bias-1,tab:gender-neutral-bias-2} show the evaluation results for 104 gender-neutral words.

\begin{table*}[t]
\small
\begin{adjustbox}{width=\linewidth,center}
\begin{tabular}{@{}l|ccc|c|llll@{}}
\toprule
\multirow{2}{*}{NOUN}   & \multicolumn{4}{c|}{Relative Gender Preference} & \multicolumn{4}{c}{Bias Class} \\ \cline{2-9}
    & \textsc{BERT} base &  \textsc{BERT} large & \textsc{ELECTRA}  & Avg.  & \textsc{BERT} base &  \textsc{BERT} large   & \textsc{ELECTRA}  & Annotated \\ \midrule
housekeeper & -2.813 & -0.573 & -2.642 & -2.009 & \textcolor{red}{female} & \textcolor{red}{female} & \textcolor{red}{female} & \textcolor{red}{female} \\ 
nurse & -2.850 & -0.568 & -2.103 & -1.840 & \textcolor{red}{female} & \textcolor{red}{female} & \textcolor{red}{female} & \textcolor{red}{female} \\ 
receptionist & -1.728 & -0.776 & -2.302 & -1.602 & \textcolor{red}{female} & \textcolor{red}{female} & \textcolor{red}{female} & \textcolor{red}{female} \\ 
hairdresser & -0.400 & -0.228 & -0.785 & -0.471 & \textcolor{red}{female} & \textcolor{red}{female} & \textcolor{red}{female} & \textcolor{red}{female} \\ 
librarian & 0.019 & -0.088 & -0.768 & -0.279 & neutral & \textcolor{red}{female} & \textcolor{red}{female} & \textcolor{red}{female} \\ 
assistant & -0.477 & 0.020 & -0.117 & -0.192 & \textcolor{red}{female} & neutral & neutral & \textcolor{red}{female} \\ 
secretary & -0.564 & 0.024 & -0.027 & -0.189 & \textcolor{red}{female} & neutral & neutral & \textcolor{red}{female} \\ 
victim & -0.075 & 0.091 & -0.323 & -0.102 & \textcolor{red}{female} & neutral & \textcolor{red}{female} & neutral \\ 
teacher & 0.129 & 0.175 & -0.595 & -0.097 & neutral & neutral & \textcolor{red}{female} & \textcolor{red}{female} \\ 
therapist & 0.002 & 0.016 & -0.233 & -0.072 & neutral & neutral & \textcolor{red}{female} & neutral \\ 
child & -0.100 & 0.073 & -0.154 & -0.060 & \textcolor{red}{female} & neutral & \textcolor{red}{female} & neutral \\ 
salesperson & -0.680 & -0.206 & 0.719 & -0.056 & \textcolor{red}{female} & \textcolor{red}{female} & \textcolor{blue}{male} & \textcolor{blue}{male} \\ 
practitioner & 0.150 & 0.361 & -0.621 & -0.037 & neutral & neutral & \textcolor{red}{female} & neutral \\ 
client & -0.157 & 0.250 & -0.165 & -0.024 & \textcolor{red}{female} & neutral & \textcolor{red}{female} & neutral \\ 
dietitian & 0.175 & 0.003 & -0.143 & 0.012 & neutral & neutral & \textcolor{red}{female} & neutral \\ 
cook & -0.150 & 0.141 & 0.048 & 0.013 & \textcolor{red}{female} & neutral & neutral & \textcolor{blue}{male} \\ 
educator & 0.278 & 0.144 & -0.375 & 0.015 & neutral & neutral & \textcolor{red}{female} & neutral \\ 
cashier & 0.009 & 0.041 & 0.017 & 0.023 & neutral & neutral & neutral & \textcolor{red}{female} \\ 
customer & -0.401 & 0.328 & 0.142 & 0.023 & \textcolor{red}{female} & neutral & neutral & neutral \\ 
attendant & -0.157 & 0.226 & 0.010 & 0.027 & \textcolor{red}{female} & neutral & neutral & \textcolor{red}{female} \\ 
designer & 0.200 & 0.173 & -0.232 & 0.047 & neutral & neutral & \textcolor{red}{female} & \textcolor{red}{female} \\ 
cleaner & 0.151 & 0.099 & -0.089 & 0.053 & neutral & neutral & neutral & \textcolor{red}{female} \\ 
teenager & 0.343 & 0.088 & -0.210 & 0.074 & neutral & neutral & \textcolor{red}{female} & neutral \\ 
passenger & 0.015 & 0.151 & 0.100 & 0.089 & neutral & neutral & neutral & neutral \\ 
guest & 0.162 & 0.258 & -0.150 & 0.090 & neutral & neutral & \textcolor{red}{female} & neutral \\ 
someone & 0.026 & 0.275 & 0.082 & 0.128 & neutral & neutral & neutral & neutral \\ 
student & 0.307 & 0.281 & -0.195 & 0.131 & neutral & neutral & \textcolor{red}{female} & neutral \\ 
clerk & 0.107 & 0.216 & 0.105 & 0.143 & neutral & neutral & neutral & \textcolor{red}{female} \\ 
visitor & 0.471 & 0.273 & -0.280 & 0.155 & neutral & neutral & \textcolor{red}{female} & neutral \\ 
counselor & 0.304 & 0.165 & 0.009 & 0.159 & neutral & neutral & neutral & \textcolor{red}{female} \\ 
editor & 0.244 & 0.161 & 0.081 & 0.162 & neutral & neutral & neutral & \textcolor{red}{female} \\ 
resident & 0.528 & 0.300 & -0.304 & 0.174 & neutral & neutral & \textcolor{red}{female} & neutral \\ 
patient & 0.009 & 0.305 & 0.217 & 0.177 & neutral & neutral & neutral & neutral \\ 
homeowner & 0.422 & 0.158 & -0.002 & 0.192 & neutral & neutral & neutral & neutral \\ 
advisee & 0.175 & 0.252 & 0.168 & 0.199 & neutral & neutral & neutral & neutral \\ 
psychologist & 0.259 & 0.232 & 0.124 & 0.205 & neutral & neutral & neutral & neutral \\ 
nutritionist & 0.474 & 0.134 & 0.020 & 0.210 & neutral & neutral & neutral & neutral \\ 
dispatcher & 0.250 & 0.118 & 0.284 & 0.217 & neutral & neutral & neutral & neutral \\ 
tailor & 0.572 & 0.382 & -0.250 & 0.235 & neutral & \textcolor{blue}{male} & \textcolor{red}{female} & \textcolor{red}{female} \\ 
employee & 0.124 & 0.228 & 0.371 & 0.241 & neutral & neutral & neutral & neutral \\ 
owner & 0.044 & 0.213 & 0.493 & 0.250 & neutral & neutral & neutral & neutral \\ 
advisor & 0.339 & 0.271 & 0.148 & 0.253 & neutral & neutral & neutral & neutral \\ 
witness & 0.287 & 0.319 & 0.187 & 0.264 & neutral & neutral & neutral & neutral \\ 
writer & 0.497 & 0.237 & 0.060 & 0.265 & neutral & neutral & neutral & \textcolor{red}{female} \\ 
undergraduate & 0.575 & 0.148 & 0.075 & 0.266 & neutral & neutral & neutral & neutral \\ 
veterinarian & 0.616 & 0.007 & 0.209 & 0.278 & neutral & neutral & neutral & neutral \\ 
pedestrian & 0.446 & 0.226 & 0.170 & 0.281 & neutral & neutral & neutral & neutral \\ 
investigator & 0.518 & 0.228 & 0.120 & 0.289 & neutral & neutral & neutral & neutral \\ 
hygienist & 0.665 & 0.274 & -0.040 & 0.300 & neutral & neutral & neutral & neutral \\ 
buyer & 0.529 & 0.190 & 0.183 & 0.300 & neutral & neutral & neutral & neutral \\ 
supervisor & 0.257 & 0.228 & 0.426 & 0.304 & neutral & neutral & neutral & \textcolor{blue}{male} \\ 
worker & 0.151 & 0.267 & 0.511 & 0.310 & neutral & neutral & neutral & neutral \\ 
bystander & 0.786 & 0.117 & 0.072 & 0.325 & \textcolor{blue}{male} & neutral & neutral & neutral \\ 
\bottomrule
\end{tabular}
\end{adjustbox}
\caption{List of gender-neutral nouns with their evaluated bias $RGP$. Female and male bias classes are assigned for 20 lowest negative and 20 highest positive $RGP$ values. Annotated bias from \citet{zhao-etal-2018-gender}. Part 1 of 2.}
\label{tab:gender-neutral-bias-1}
\end{table*}

\begin{table*}[t]
\small
\begin{adjustbox}{width=\linewidth,center}
\begin{tabular}{@{}l|ccc|c|llll@{}}
\toprule
\multirow{2}{*}{NOUN}   & \multicolumn{4}{c|}{Relative Gender Preference} & \multicolumn{4}{c}{Bias Class} \\ \cline{2-9}
    & \textsc{BERT} base &  \textsc{BERT} large & \textsc{ELECTRA}  & Avg.  & \textsc{BERT} base &  \textsc{BERT} large   & \textsc{ELECTRA}  & Annotated \\ \midrule
chemist & 0.579 & 0.311 & 0.107 & 0.332 & neutral & neutral & neutral & neutral \\ 
administrator & 0.428 & 0.236 & 0.350 & 0.338 & neutral & neutral & neutral & neutral \\ 
examiner & 0.445 & 0.281 & 0.296 & 0.341 & neutral & neutral & neutral & neutral \\ 
broker & 0.376 & 0.358 & 0.295 & 0.343 & neutral & neutral & neutral & neutral \\ 
instructor & 0.413 & 0.196 & 0.436 & 0.348 & neutral & neutral & neutral & neutral \\ 
developer & 0.536 & 0.338 & 0.172 & 0.349 & neutral & neutral & neutral & \textcolor{blue}{male} \\ 
technician & 0.312 & 0.362 & 0.400 & 0.358 & neutral & neutral & neutral & neutral \\ 
baker & 0.622 & 0.287 & 0.178 & 0.362 & neutral & neutral & neutral & \textcolor{red}{female} \\ 
planner & 0.611 & 0.341 & 0.147 & 0.366 & neutral & neutral & neutral & neutral \\ 
bartender & 0.628 & 0.282 & 0.293 & 0.401 & neutral & neutral & neutral & neutral \\ 
paramedic & 0.787 & 0.094 & 0.333 & 0.405 & \textcolor{blue}{male} & neutral & neutral & neutral \\ 
protester & 0.722 & 0.498 & 0.019 & 0.413 & neutral & \textcolor{blue}{male} & neutral & neutral \\ 
specialist & 0.501 & 0.363 & 0.392 & 0.419 & neutral & \textcolor{blue}{male} & neutral & neutral \\ 
electrician & 0.935 & 0.283 & 0.076 & 0.431 & \textcolor{blue}{male} & neutral & neutral & neutral \\ 
physician & 0.438 & 0.359 & 0.502 & 0.433 & neutral & neutral & neutral & \textcolor{blue}{male} \\ 
pathologist & 0.817 & 0.307 & 0.181 & 0.435 & \textcolor{blue}{male} & neutral & neutral & neutral \\ 
analyst & 0.645 & 0.315 & 0.361 & 0.440 & neutral & neutral & neutral & \textcolor{blue}{male} \\ 
appraiser & 0.729 & 0.305 & 0.302 & 0.445 & neutral & neutral & neutral & neutral \\ 
onlooker & 0.978 & 0.093 & 0.274 & 0.448 & \textcolor{blue}{male} & neutral & neutral & neutral \\ 
janitor & 0.702 & 0.493 & 0.174 & 0.456 & neutral & \textcolor{blue}{male} & neutral & \textcolor{blue}{male} \\ 
mover & 0.717 & 0.407 & 0.253 & 0.459 & neutral & \textcolor{blue}{male} & neutral & \textcolor{blue}{male} \\ 
chef & 0.682 & 0.348 & 0.352 & 0.460 & neutral & neutral & neutral & neutral \\ 
lawyer & 0.696 & 0.271 & 0.421 & 0.462 & neutral & neutral & neutral & \textcolor{blue}{male} \\ 
paralegal & 0.829 & 0.247 & 0.313 & 0.463 & \textcolor{blue}{male} & neutral & neutral & neutral \\ 
doctor & 0.723 & 0.355 & 0.322 & 0.467 & neutral & neutral & neutral & neutral \\ 
auditor & 0.654 & 0.329 & 0.504 & 0.496 & neutral & neutral & neutral & \textcolor{red}{female} \\ 
officer & 0.465 & 0.463 & 0.584 & 0.504 & neutral & \textcolor{blue}{male} & \textcolor{blue}{male} & neutral \\ 
surgeon & 0.368 & 0.417 & 0.733 & 0.506 & neutral & \textcolor{blue}{male} & \textcolor{blue}{male} & neutral \\ 
programmer & 0.543 & 0.304 & 0.684 & 0.510 & neutral & neutral & \textcolor{blue}{male} & neutral \\ 
scientist & 0.568 & 0.427 & 0.548 & 0.514 & neutral & \textcolor{blue}{male} & neutral & neutral \\ 
painter & 0.721 & 0.298 & 0.555 & 0.525 & neutral & neutral & \textcolor{blue}{male} & neutral \\ 
pharmacist & 0.862 & 0.244 & 0.495 & 0.534 & \textcolor{blue}{male} & neutral & neutral & neutral \\ 
laborer & 0.996 & 0.557 & 0.058 & 0.537 & \textcolor{blue}{male} & \textcolor{blue}{male} & neutral & \textcolor{blue}{male} \\ 
machinist & 0.821 & 0.449 & 0.361 & 0.544 & \textcolor{blue}{male} & \textcolor{blue}{male} & neutral & neutral \\ 
architect & 0.790 & 0.243 & 0.609 & 0.547 & \textcolor{blue}{male} & neutral & \textcolor{blue}{male} & neutral \\ 
taxpayer & 0.785 & 0.525 & 0.339 & 0.550 & \textcolor{blue}{male} & \textcolor{blue}{male} & neutral & neutral \\ 
chief & 0.595 & 0.472 & 0.628 & 0.565 & neutral & \textcolor{blue}{male} & \textcolor{blue}{male} & \textcolor{blue}{male} \\ 
inspector & 0.631 & 0.344 & 0.726 & 0.567 & neutral & neutral & \textcolor{blue}{male} & neutral \\ 
plumber & 1.186 & 0.468 & 0.205 & 0.620 & \textcolor{blue}{male} & \textcolor{blue}{male} & neutral & neutral \\ 
construction worker & 0.770 & 0.326 & 0.769 & 0.622 & \textcolor{blue}{male} & neutral & \textcolor{blue}{male} & \textcolor{blue}{male} \\ 
driver & 0.847 & 0.415 & 0.603 & 0.622 & \textcolor{blue}{male} & \textcolor{blue}{male} & \textcolor{blue}{male} & \textcolor{blue}{male} \\ 
manager & 0.456 & 0.346 & 1.084 & 0.628 & neutral & neutral & \textcolor{blue}{male} & \textcolor{blue}{male} \\ 
engineer & 0.562 & 0.385 & 0.987 & 0.645 & neutral & \textcolor{blue}{male} & \textcolor{blue}{male} & neutral \\ 
sheriff & 0.850 & 0.396 & 0.708 & 0.651 & \textcolor{blue}{male} & \textcolor{blue}{male} & \textcolor{blue}{male} & \textcolor{blue}{male} \\ 
CEO & 0.701 & 0.353 & 0.989 & 0.681 & neutral & neutral & \textcolor{blue}{male} & \textcolor{blue}{male} \\ 
mechanic & 0.752 & 0.307 & 1.098 & 0.719 & \textcolor{blue}{male} & neutral & \textcolor{blue}{male} & \textcolor{blue}{male} \\ 
guard & 0.907 & 0.586 & 0.720 & 0.738 & \textcolor{blue}{male} & \textcolor{blue}{male} & \textcolor{blue}{male} & \textcolor{blue}{male} \\ 
accountant & 0.610 & 0.291 & 1.350 & 0.750 & neutral & neutral & \textcolor{blue}{male} & \textcolor{red}{female} \\ 
farmer & 1.044 & 0.477 & 0.736 & 0.753 & \textcolor{blue}{male} & \textcolor{blue}{male} & \textcolor{blue}{male} & \textcolor{blue}{male} \\ 
firefighter & 1.294 & 0.438 & 0.604 & 0.779 & \textcolor{blue}{male} & \textcolor{blue}{male} & \textcolor{blue}{male} & neutral \\ 
carpenter & 0.934 & 0.415 & 1.263 & 0.870 & \textcolor{blue}{male} & \textcolor{blue}{male} & \textcolor{blue}{male} & \textcolor{blue}{male} \\
\bottomrule
\end{tabular}
\end{adjustbox}
\caption{List of gender-neutral nouns with their evaluated bias $RGP$. Female and male bias classes are assigned for 20 lowest negative and 20 highest positive $RGP$ values. Annotated bias from \citet{zhao-etal-2018-gender}. Part 2 of 2.}
\label{tab:gender-neutral-bias-2}
\end{table*}

\end{document}